\documentclass[pmlr]{jmlr}

\usepackage{longtable}

\usepackage{booktabs}
\usepackage[load-configurations=version-1]{siunitx} 

\makeatletter
\def\set@curr@file#1{\def\@curr@file{#1}} 
\makeatother

\theorembodyfont{\upshape}
\theoremheaderfont{\scshape}
\theorempostheader{:}
\theoremsep{\newline}

\usepackage{enumitem}
\usepackage{algorithm}
\usepackage{graphics}
\usepackage{algorithmic}
\usepackage{multirow}
\usepackage{adjustbox}

\title[]{Learning by Teaching, with Application to Neural Architecture Search}

\author{\Name{Parth Sheth
}
       \Email{parthfour@gmail.com} 
       \AND
       \Name{Yueyu Jiang}
       \Email{y5jiang@eng.ucsd.edu} 
       \AND
       \Name{Pengtao Xie\textsuperscript{*}}
       \Email{p1xie@eng.ucsd.edu}\\
       \addr 
University of California San Diego
\AND
       }

\begin{document}

\maketitle

\begin{abstract}
In human learning, an effective skill in improving learning outcomes is learning by teaching: a learner deepens his/her understanding of a topic by teaching this topic to others. In this paper, we aim to borrow this teaching-driven learning methodology from humans and leverage it to train more performant machine learning models, by proposing a novel ML framework referred to as learning by teaching (LBT). In the LBT framework, a teacher model improves itself by teaching a student model to learn well. Specifically, the teacher creates a pseudo-labeled dataset and uses it to train a student model. Based on how the student performs on a validation dataset, the teacher re-learns its model and re-teaches the student until the student achieves great validation performance. Our framework is based on three-level optimization which contains three stages:  teacher learns; teacher teaches student; teacher re-learns based on how well the student performs. A simple but efficient algorithm is developed to solve the three-level optimization problem. We apply LBT to search neural architectures on CIFAR-10, CIFAR-100, and ImageNet. The efficacy of our method is demonstrated in various experiments.
\end{abstract}

\section{Introduction}

\let\thefootnote\relax\footnotetext{$^*$Corresponding author.}

As the saying goes, a good learner is also a good teacher. In human learning, a commonly adopted  strategy is to learn by teaching  others. In the process of explaining a topic to others, the learner can further enhance his/her understanding of this topic. The efficacy of teaching in helping improve learning has been demonstrated in many studies. In~\citep{fiorella2013relative}, the studies showed that students who teach what they have learned to their peers achieve better understanding and knowledge retention than students spending the same time re-studying. In~\citep{koh2018learning}, the study shows that teaching improves the teacher’s learning because it encourages the teacher to retrieve what they have previously learned.

\begin{figure}[t]
    \centering
 \includegraphics[width=0.5\columnwidth]{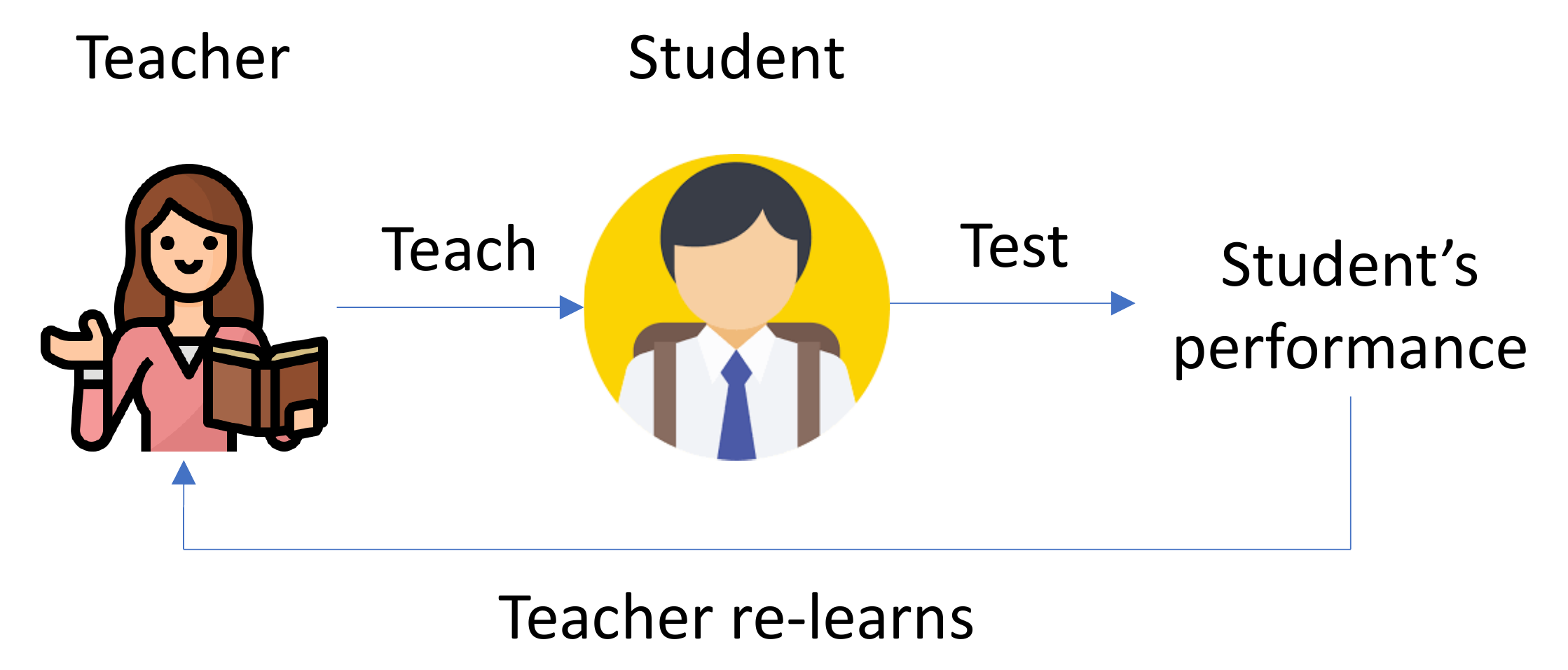}
       \caption{Illustration of learning by teaching. The teacher first learns a topic. Then the teacher teaches this topic to the student and the student learns this topic. The student performs a test to check how well he/she masters this topic. Based on the student's performance on the test, the teacher re-learns this topic to improve his/her understanding.}
 \label{fig:illus}
\end{figure}

This teaching-driven learning methodology of humans motivates us to think about whether it can benefit machine learning as well. Toward this goal, we propose a novel ML framework called learning by teaching (LBT) (as illustrated in Figure~\ref{fig:illus}), which improves the learning outcome of a model by encouraging this model to teach other models to perform well. 
 In our framework, there is a teacher model and a student model, which perform the same target task (e.g., text classification, time-series forecasting, etc.). The eventual goal is to make the teacher perform well on the target task. The way to achieve that is to let the teacher teach the student and use the student's performance as a feedback to guide the teacher to improve its learning capability and outcome. The teacher model consists of a learnable architecture and a set of learnable network weights. The student model consists of a predefined  architecture (by human experts) and a set of learnable network weights. Similar to~\citep{HintonVD15}, teaching is conducted via pseudo-labeling: given an unlabeled dataset $U$, the teacher uses its intermediately trained model to make predictions on the input data examples in $U$; then the student model is trained on these pseudo-labeled data examples. Teacher-student learning based on pseudo-labeling has been studied in many previous works~\citep{papernot2016semi,tarvainen2017mean,xie2020self}. Our work differs from previous ones in that we aim to improve the learning ability of the teacher by letting it teach a student while previous works focus on improving the learning ability of a student model by letting it be taught by a fixed teacher model. In other words, our work focuses on learning the teacher while previous works focus on learning the student.

In our framework, the learning of the teacher and student are organized into three stages. 
 In the first stage, the teacher learns its network weights on a training dataset while temporarily fixing its architecture. 
 In the second stage, the teacher performs pseudo-labeling on an unlabeled dataset and uses the pseudo-labeled dataset to train the student model. Specifically, the teacher applies its model trained in the first stage to make predictions on unlabeled data examples and the student model is trained to fit these predictions. In the third stage, the student's model is evaluated on a validation set and the teacher adjusts its architecture based on the student's validation performance. The three stages are organized into a three-level optimization framework and are performed end-to-end in a unified manner, where earlier stages influence later stages and vice versa. 
 We apply LBT for neural architecture search. Experiments on CIFAR-100, CIFAR-10, and ImageNet~\citep{deng2009imagenet} demonstrate the effectiveness of our method.

The major contributions of this paper include:
\begin{itemize}
\item Motivated by the teaching-driven learning methodology of humans, we develop a novel machine learning framework called learning by teaching (LBT). In our approach, a teacher creates a pseudo-labeled dataset and uses it to train a student model. Based on how the student performs on the validation dataset, the teacher re-learns its model and re-teaches the student until the student achieves great validation performance. 
\item To formulate LBT, we  develop a three-level optimization framework. This framework consists of three learning stages: 1) teacher performs learning; 2) teacher teaches what it has learned to a student; 3) teacher re-learns based on the performance of the student. 
\item An efficient algorithm is developed to solve the three-level optimization problem. 
\item We apply LBT for neural architecture search on CIFAR-100, CIFAR-10, and ImageNet, where the results show that our method is very effective in searching highly-performing neural architectures. 
\end{itemize}

\section{Related Works}
\subsection{Neural Architecture Search}
Neural architecture search (NAS) is the task of developing algorithms to automatically find out architectures that can yield high ML-performance. Existing NAS methods can be categorized into three groups. Methods in the first group~\citep{zoph2016neural,pham2018efficient,zoph2018learning} are based on reinforcement learning,  where an architecture generation policy is learned by maximizing ML performance on validation data. Methods in the second group~\citep{cai2018proxylessnas,liu2018darts,xie2018snas} are gradient-based and differentiable. These methods adopt a network pruning strategy where an overparameterized network with many building blocks is pruned into the final architecture and the optimal pruning is achieved by minimizing the validation loss. Methods in the third group~\citep{liu2017hierarchical,real2019regularized} are based on evolutionary algorithms where architectures are represented as a population. Highly-performing architectures are allowed to generate offspring while poorly-performing architectures are eliminated.

\subsection{Teacher-Student Learning}
Teacher-student learning has been investigated in knowledge distillation~\citep{hinton2015distilling}, adversarial robustness~\citep{carlini2017towards}, self-supervised learning~\citep{xie2020self}, etc. Most of these methods are based on pseudo-labeling. In these existing methods, the focus is to learn a student model with the help of a trained and fixed teacher model. In these works, the teacher model is not updated. In contrast, our method focuses on learning a teacher model, by letting it teach a student model. The teacher model constantly updates itself based on the teaching outcome. Teacher-student learning has been investigated in several neural architecture search works as well~\citep{LiPYWLLC20,abs-2006-08341,GuT20}. In these works, when searching the architecture of a  student model, pseudo-labels generated by a trained teacher model whose architecture is fixed are leveraged. Our work differs from these works in that we focus on searching the architecture of a teacher model by letting it teach a student model where the student's architecture is fixed, whereas the existing works focus on searching the architecture of a student model by letting it be taught by a teacher where the teacher's architecture is fixed. 
 In a recent work~\citep{pham2020meta} which was conducted independently of and in parallel to our work, the teacher model is updated based on student's performance. Our work differs from this one in that our work is based on a three-level optimization framework which searches teacher's architecture by minimizing student's validation loss and trains teacher's network weights before using teacher to generate pseudo-labels, whereas in~\citep{pham2020meta} the framework is based on two-level optimization which has no architecture search and does not train the teacher before using it to perform pseudo-labeling. In~\citep{abs-1912-07768}, a meta-learning method is developed to learn a deep generative model which generates synthetic labeled-data. A student model leverages the synthesized data to search its architecture. Our work differs from this method in that we focus on searching the teacher's architecture via three-level optimization while \citep{abs-1912-07768} focuses on searching the student's architecture via meta-learning.

\section{Methods}
\begin{figure}[t]
    \centering
 \includegraphics[width=0.5\columnwidth]{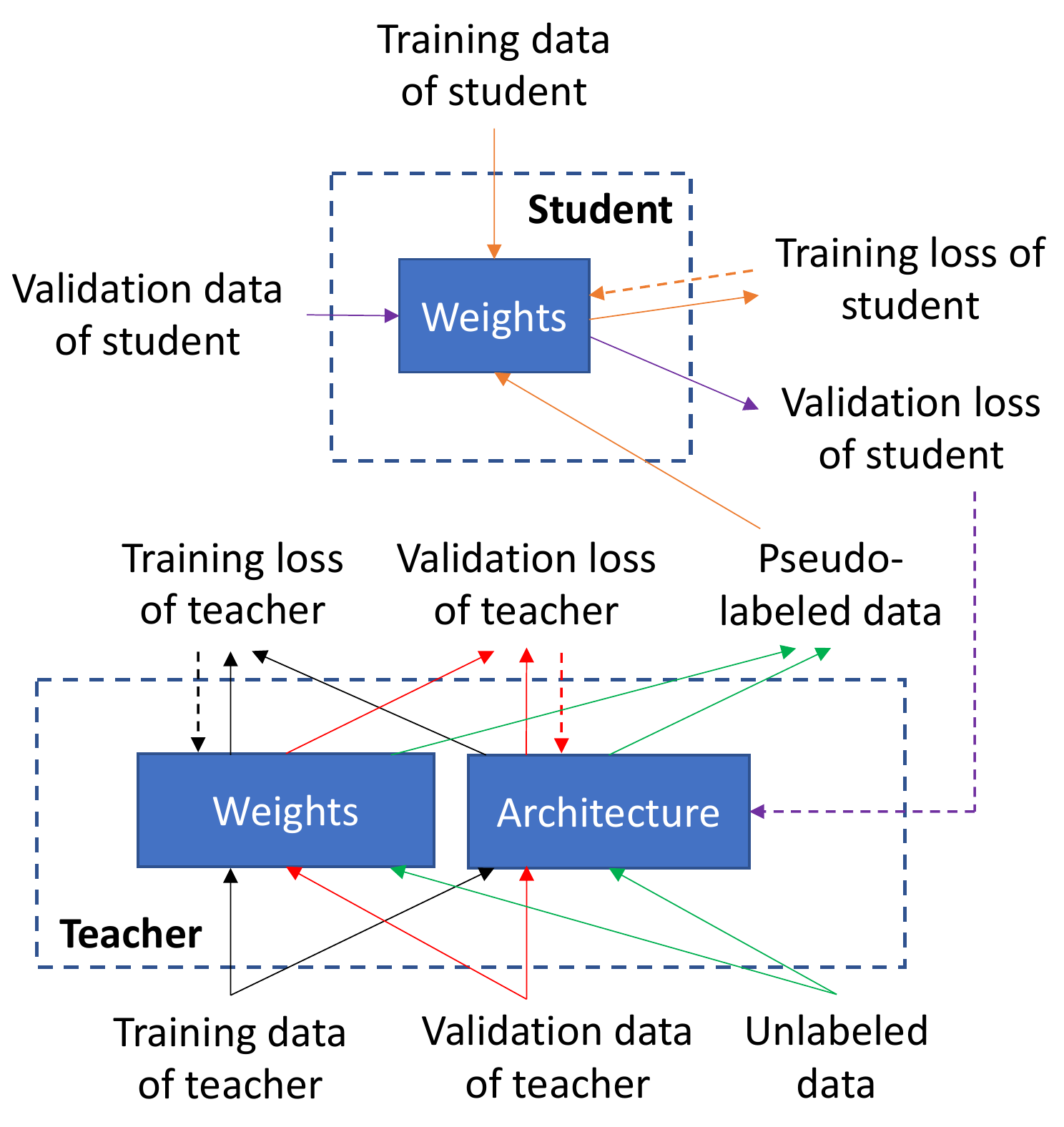}
       \caption{Learning by teaching. The solid arrows denote a forward pass where predictions are made and training/validation losses are defined. 
  The dotted arrows denote a backward process 
  where gradients of losses are calculated and parameters are updated.}
 \label{fig:arch}
\end{figure}

In this section, we propose a three-level optimization framework to formalize learning-by-teaching (LBT) (as shown in Figure~\ref{fig:arch}) and develop an efficient optimization algorithm for solving the LBT problem.

\subsection{Learning by Teaching}
\begin{table}[t]
\caption{Notations in Learning by Teaching}
\centering
\begin{tabular}{l|l}
\hline
Notation & Meaning \\
\hline
$A$ & Architecture of the teacher\\
$T$ & Network weights of the teacher\\
$S$ & Network weights of the student\\
$D_{t}^{(\textrm{tr})}$ & Training data of the teacher\\
$D_{t}^{(\textrm{val})}$ & Validation data of the teacher\\
$D_{s}^{(\textrm{tr})}$ & Training data of the student\\
$D_{s}^{(\textrm{val})}$ & Validation data of the student\\
$D_u$ & Unlabeled dataset \\
\hline
\end{tabular}
\label{tb:notations}
\end{table}

In our framework, there is a teacher model and a student model, which both study how to perform the same target task. Without loss of generality, we assume the target task is classification. The eventual goal is to make the teacher achieve better learning outcomes. The way to achieve this goal to let the teacher teach the student to perform well on the target task. The intuition behind LBT is that a teacher needs to learn a topic very well in order to  teach this topic to a student clearly. Teaching is performed based on pseudo-labeling~\citep{hinton2015distilling}: the teacher uses its model to generate a pseudo-labeled dataset; the student is trained on the pseudo-labeled dataset. 
The teacher has a learnable neural architecture $A$ and a set of learnable  network weights $T$. 
The student has a predefined architecture (by humans) and a set of learnable network weights $S$. The teacher has a training dataset $D^{(\textrm{tr})}_t$ and a validation dataset $D^{(\textrm{val})}_t$. The student has a training dataset $D^{(\textrm{tr})}_s$ and a validation dataset $D^{(\textrm{val})}_s$. There is an unlabeled dataset $D_u$ where pseudo labeling is performed. In our framework, both the teacher and student perform learning, which is organized into three stages. In the first stage, the teacher fixes its architecture and trains its network weights by minimizing the training loss defined on $D^{(\textrm{tr})}_t$:
\begin{equation}
    T^*(A) =\textrm{min}_{T} \; L(T,A,D_t^{(\textrm{tr})}). 
\end{equation}

The architecture $A$ is needed to calculate the loss on training examples. However, it cannot be updated by minimizing the training loss. Otherwise, a degenerated solution  will be produced where $A$ has very large capacity to overfit the training examples but will yield poor prediction outcomes on unseen examples. $T^*(A)$ is a function of $A$: a different $A$ will result in a different training loss $L(A, T,D_t^{(\textrm{tr})})$; $T$ trained by minimizing $L(A, T,D_t^{(\textrm{tr})})$ will be different as well. In the second stage, the teacher teaches a student via pseudo-labeling. Given an unlabeled dataset $D_u=\{x_i\}_{i=1}^N$, the teacher uses its model $T^*(A)$ trained in the first stage to make predictions on $D_u$. Assuming the task is classification with $K$ classes, the prediction $f(x_i;T^*(A))$ on $x_i$ would be a $K$-dimensional vector, where the $k$-th element indicates the probability that $x_i$ belongs to the $k$-th class and the sum of elements in $f(x_i;T^*(A))$ is one. Let $D_{pl}(D_u,T^*(A))=\{(x_i,f(x_i;T^*(A)))\}_{i=1}^N$ denote the pseudo-labeled dataset. The network weights $S$ of the student is trained on $D_{pl}(D_u,T^*(A))$ and a human-labeled training set $D_{s}^{(\textrm{tr})}$: 
\begin{equation*}
    S^*(T^*(A)) = \textrm{min}_S \;  L(S, D_s^{(\textrm{tr})} )+\lambda  L(S, D_{pl}(D_u,T^*(A))).
\end{equation*}
where $L(\cdot)$ denotes a cross-entropy loss and $\lambda$ is a tradeoff parameter. 
$S^*(T^*(A))$ is a function of $T^*(A)$: a different $T^*(A)$ will result in a different pseudo-labeled dataset $D_{pl}(D_u,T^*(A))$ which will render the training loss to be different; a different training loss will result in a different $S^*(T^*(A))$. 
In the third stage, the student's model $S^*(T^*(A))$ trained in the second stage is validated on $D_{s}^{(\textrm{val})}$. Besides, we also validate the teacher's model $T^*(A)$ trained in the first stage on $D_{t}^{(\textrm{val})}$. The validation performances provide feedback on how good the teacher's architecture $A$ is. At this stage, $A$ is optimized by minimizing the validation losses:
\begin{equation}
    \textrm{min}_{A}  
    \; L(T^*(A),A,  D_t^{(\textrm{val})})  + \gamma L(S^*(T^*(A)), D_s^{(\textrm{val})}),
\end{equation}
where $\gamma$ is a tradeoff parameter.

Given the three learning stages, we propose a three-level optimization framework to stitch them together:
\begin{equation}
\begin{array}{l}
\underset{A}{\textrm{min}}
  \;\;  L(T^*(A),A, D_t^{(\textrm{val})})) + \gamma  L(S^*(T^*(A)), D_s^{(\textrm{val})})\\
      s.t. \;\;\; S^*(T^*(A)) =
      \underset{S}{\textrm{min}}
    \;\;  L(S, D_s^{(\textrm{tr})} )+\lambda  L(S, D_{pl}(D_u,T^*(A)))\\
    \quad\;\;\;\; T^*(A) =
    \underset{T}{\textrm{min}} \; L(A, T,D_t^{(\textrm{tr})})
\end{array}
\label{eq:lbt}
\end{equation}
From bottom to top, the three optimization problems correspond to the first, second, and third stage respectively. The first two optimization problems are on the constraints of the third optimization problem. The three stages are performed end-to-end in a joint manner where different stages mutually influence each other. $T^*(A)$ trained in the first stage is used to generate pseudo-labeled dataset in the second stage; $T^*(A)$ and $S^*(T^*(A))$ trained in the first two stages are validated in the third stage; after $A$ is updated in the third stage, it will render the training loss in the first stage to be changed, which accordingly results in a new $T^*(A)$. For computational efficiency, we search $A$ in a differentiable way~\citep{liu2018darts}: given an overparameterized network, a subnetwork is carved out as the final architecture. The overparameterized network contains a large number of basic building blocks such as convolution operations, pooling operations, etc. The output of each building block is multiplied with a scalar. The search algorithm optimizes these scalars by minimizing validation losses. In the end, building blocks with largest scalars form the final architecture.

\subsection{Optimization Algorithm}

In this section, we develop a gradient-based  algorithm to solve the learning by teaching (LBT) problem. Drawing insights from~\citep{liu2018darts}, we calculate the gradient of $L(A, T,D_t^{(\textrm{tr})})$ w.r.t $T$, update $T$ by one-step gradient descent and get $T'$, which is used as an approximation of $T^{*}(A)$. 
We substitute this approximation  into $L(S, D_s^{(\textrm{tr})} )+\lambda  L(S, D_{pl}(D_u,T^*(A)))$, yielding an approximated objective $O_s$. Similarly,  we approximate  $S^*(T^{*}(A))$ using one-step gradient descent update of  $S$ using the gradient of $O_s$. Lastly, we substitute the approximations of $T^{*}(A)$ and $S^*(T^{*}(A))$ into $ L(T^*(A), D_t^{(\textrm{val})}))+ \gamma  L(S^*(T^*(A)), D_s^{(\textrm{val})})$ and  perform gradient-descent update of  $A$ by minimizing the approximated validation loss. Let $\nabla^2_{Y,X}f(X,Y)$ denote $\frac{\partial f(X,Y)}{\partial X\partial Y}$.

First, we approximate $T^{*}(A)$ using 
\begin{equation}
    T'=T - \xi_{t}  \nabla_{T}L(T, A, D_{t}^{(\mathrm{tr})})
    \label{eq:update_t}
\end{equation}
where $\xi_{t}$ is a learning rate. Substituting   $T'$ into $L(S, D_s^{(\textrm{tr})} )+\lambda  L(S, D_{pl}(D_u,T^*(A)))$ results in an approximated objective $O_s=L(S, D_s^{(\textrm{tr})} )+\lambda  L(S, D_{pl}(D_u,T'))$. Next,  we approximate  $S^*(T^{*}(A))$ using one-step gradient descent update of  $S$ w.r.t $O_s$:
\begin{equation}
    S'=S - \xi_{s}  \nabla_{S}(L(S, D_s^{(\textrm{tr})} )+\lambda  L(S, D_{pl}(D_u,T'))).
    \label{eq:update_s}
\end{equation}
Finally, we plug  $T'$ and $S'$ into $ L(T^*(A), D_t^{(\textrm{val})}))+\gamma L(S^*(T^*(A)), D_s^{(\textrm{val})})$ and get $O_v=L(T', D_t^{(\textrm{val})})+\gamma L(S', D_s^{(\textrm{val})})$. We can update the teacher's architecture $A$ by descending the gradient of $O_v$ w.r.t $A$:
\begin{equation}
\begin{array}{l}
A\gets A-\eta ( \nabla_A L(T', A, D_t^{(\textrm{val})})+\gamma \nabla_A L(S', D_s^{(\textrm{val})}))   
    \end{array}
    \label{eq:update_a}
\end{equation}
where 
\begin{equation}
\begin{array}{l}
     \nabla_{A} L(T',A,  D_t^{(\textrm{val})}) =  \\
     \nabla_{A}  L(T - \xi_{t}  \nabla_{T}L(T, A, D_{t}^{(\mathrm{tr})}),A,  D_t^{(\textrm{val})})=\\
     - \xi_{t}  \nabla^2_{A,T}L(T, A, D_{t}^{(\mathrm{tr})})\nabla_{T'} L(T',A,  D_t^{(\textrm{val})})+ \nabla_{A} L(T', A, D_t^{(\textrm{val})})
\end{array}
\label{eq:descent_arch}
\end{equation}

The matrix-vector multiplication in the first term on the third line is computationally expensive. To reduce computational cost, following~\citep{liu2018darts}, we approximate the multiplication using a finite difference:
\begin{equation}
\begin{array}{ll}
     \nabla_{A, T}^{2} L(T,A, D_{t}^{(\mathrm{tr})})\nabla_{T'} L(T',A,D_t^{(\textrm{val})})\approx 
     \frac{1}{2\alpha}(\nabla_{A} L( T^{+},A, D_{t}^{(\mathrm{tr})})-\nabla_{A} L(T^{-},A, D_{t}^{(\mathrm{tr})})),
\end{array}
\label{eq:finite-aw}
\end{equation}
where $T^{\pm}=T \pm \alpha \nabla_{T^{\prime}} L(T',A, D_t^{(\textrm{val})})$ and $\alpha$ is  $0.01 /\|\nabla_{T'} L(T',A,D_t^{(\textrm{val})})\|_{2}$.

For $\nabla_A L(S', D_s^{(\textrm{val})})$ in Eq.(\ref{eq:update_a}), it can be calculated as $\frac{\partial S'}{\partial A} \nabla_{S'}L(S',D_s^{(\textrm{val})})$ according to the chain rule, where 
\begin{align}
\frac{\partial S'}{\partial A}&=\frac{\partial (S - \xi_{s}  \nabla_{S}(L(S, D_s^{(\textrm{tr})} )+\lambda  L(S, D_{pl}(D_u,T'))))}{\partial A}\\
& =  \frac{\partial ( - \xi_{s}\lambda   \nabla_{S} L(S, D_{pl}(D_u,T')))}{\partial A}\\
& =  - \xi_{s}\lambda \frac{\partial T' }{\partial A}\nabla^2_{T',S}L(S, D_{pl}(D_u,T'))
\end{align}
 For $\frac{\partial T' }{\partial A}$, it can be calculated as:
\begin{align}
    \frac{\partial T' }{\partial A}&=\frac{\partial (T - \xi_{t}  \nabla_{T}L(T, A, D_{t}^{(\mathrm{tr})}))}{\partial A}= - \xi_{t}   \nabla^2_{A,T}L(T, A, D_{t}^{(\mathrm{tr})})
\end{align}

The algorithm for solving LBT is presented in Algorithm~\ref{algo:algo}.
\begin{algorithm}[H]
\SetAlgoLined
 \While{not converged}{
1. Update the teacher's network weights $T$ using Eq.(\ref{eq:update_t})\\
2. Update the student's network weights $S$ using Eq.(\ref{eq:update_s})\\
3. Update the teacher's architecture $A$ using Eq.(\ref{eq:update_a})
 }
 \caption{Optimization algorithm for learning by teaching}
 \label{algo:algo}
\end{algorithm}

\section{Experiments}
In this section, we apply learning-by-teaching (LBT) to search neural architectures in image classification tasks. We follow the experimental protocol in~\citep{liu2018darts}, consisting of two phrases: one for architecture search and the other for architecture evaluation. In the search phrase, an optimal cell is searched by minimizing the validation loss. In the evaluation phrase, the searched cell is replicated and composed into a large network, which is trained from scratch on training and validation sets. Its performance is reported on the test set.  More hyperparameter settings, additional results, and significance tests of results are deferred to the supplements.
\subsection{Datasets}
The experiments were conducted on three image classification datasets:  CIFAR-10, CIFAR-100,  and ImageNet~\citep{deng2009imagenet}, with 10, 100, and 1000 classes respectively. For CIFAR-10 and CIFAR-100, each of them is split into a 25K training set, a 25K validation set, and a 5K test set. The training set is used as  $D_{t}^{\textrm{(tr)}}$ of the teacher and  $D_{s}^{\textrm{(tr)}}$ of the student. The validation set is used as  $D_{t}^{\textrm{(val)}}$ of the teacher and  $D_{s}^{\textrm{(val)}}$ of the student. For experiments on CIFAR-10, input images in CIFAR-100 (removing labels) are used as the unlabeled dataset $D_u$. For experiments on CIFAR-100, input images in CIFAR-10 (removing labels) are used as the unlabeled dataset $D_u$. For ImageNet, it is split into a training set of 1.2M images and a test set of 50K images.

\subsection{Experimental Settings}
In LBT, for the search space of $A$, we experimented the spaces in DARTS~\citep{liu2018darts}, P-DARTS~\citep{chen2019progressive}, and PC-DARTS~\citep{abs-1907-05737}. 
 These search spaces  are similar, with the following candidate operations:   $3\times 3$ and $5\times 5$ separable convolutions, $3\times 3$ and $5\times 5$ dilated separable convolutions, $3\times 3$ max pooling, $3\times 3$ average pooling, identity, and zero. For the student's architecture, we experimented with ResNet-18 and ResNet-50~\citep{resnet}. $\lambda$ and $\gamma$ are both set to 1.

During architecture search, for CIFAR-10 and CIFAR-100, the teacher's architecture is a stack of 8 cells, each consisting of 7 nodes.  The initial channel number was set to 16.  The rest hyperparameters for the teacher's architecture and network weights follow those in DARTS, P-DARTS, and PC-DARTS. The search algorithm ran for 50 epochs, with a batch size of 64. Network weights are optimized using SGD, with an initial learning rate of 0.025 (adjusted using a cosine decay scheduler), a momentum of 0.9, and a weight decay of 3e-4.
For architecture search on ImageNet, following~\citep{abs-1907-05737}, we randomly sample 10\% of the 1.2M ImageNet images as $D_t^{(\textrm{tr})}$ and $D_s^{(\textrm{tr})}$ in LBT, randomly sample 2.5\% of the 1.2M images as $D_t^{(\textrm{val})}$ and $D_s^{(\textrm{val})}$, and  randomly sample another 10\% of the 1.2M  images as $D_u$. 

During architecture evaluation, for CIFAR-10 and CIFAR-100, 20 copies of the optimal cell searched in the search phrase are stacked into a large network, which is trained using the combined training and validation datasets. The initial channel number was set to 36. The network was trained for 600 epochs, with mini-batch size set to 96. These experiments were conducted on a Tesla v100 GPU. 
For ImageNet, we evaluate the architectures searched by PC-DARTS on the subset of ImageNet and architectures searched by DARTS-2nd and P-DARTS on CIFAR-10 and CIFAR-100, by stacking 14 searched cells into a large network and training it on the 1.2M training images and reporting its performance on the 50K test images. The initial channel number was set to 48. The network was trained for 250 epochs with a batch size of 1024 on 8 Tesla v100s. Each LBT experiment was repeated for ten times with different random seeds. 
Mean and standard deviation of the 10 runs are reported.

\subsection{Results}

In Table~\ref{tab:cifar100}, we compare different NAS methods in terms of  classification error on the test set of CIFAR-100, number of network weights, and search cost measured using GPU days. From these results, we observe the following. \textbf{First}, with the help of our proposed learning-by-teaching (LBT) framework, the architectures searched by various methods including DARTS, P-DARTS, and PC-DARTS can be greatly improved. For example, applying LBT to DARTS-2nd, the error is reduced from 20.58\% to 17.06\%. With the assistance of LBT, the error of P-DARTS decreases from 17.49\% to 16.29\%. Without using LBT, the error of PC-DARTS is 17.96\%; adding LBT reduces this error to 16.88\%. These results strongly demonstrate that LBT is an effective learning framework that helps to improve a wide variety of NAS methods. In our method, the teacher model improves its learning ability by teaching a student model to perform well on the classification task. The student is trained on the pseudo-labeled dataset created by the teacher. If the student does not perform well on the validation set, that means the  pseudo labels  are not correct, which indicates  the teacher's model is not accurate. To avoid such an outcome, the teacher enforces itself to learn better to generate correct pseudo labels. \textbf{Second}, a stronger student helps the teacher to learn better. Here we consider a student model is stronger if its architecture (manually designed) is more powerful and expressive. 
\begin{table}[t]
 \caption{Results on CIFAR-100, including classification error (\%) on the test set, number of model weights (millions), and search cost (GPU days). LBT(RN18,DARTS-1st) represents that in LBT the search space is the same as that of DARTS-1st and the architecture of the student is ResNet-18. 
     Similar meanings hold for other notations in such a format. RN50 denotes ResNet-50. 
    DARTS-1st  and 
DARTS-2nd indicates that first-order and second-order approximation is used in DARTS' optimization algorithm. 
    * denotes that the results are taken from DARTS$^{-}$ \citep{abs-2009-01027}. $\dag$ denotes that we re-ran this method for 10 times.  The search cost is measured by GPU days on a Tesla v100.
    }
    \centering
    \begin{tabular}{l|ccc}
    \toprule
    Method & Error(\%)& Param(M)& Cost\\
    \midrule
    *ResNet \citep{he2016deep}&22.10&1.7&-\\
     *DenseNet \citep{HuangLMW17}&17.18&25.6 &-\\
    \hline
    *PNAS \citep{LiuZNSHLFYHM18}&19.53&3.2&150\\
    *ENAS \citep{pham2018efficient}&19.43&4.6&0.5\\
        *AmoebaNet \citep{real2019regularized}&18.93&3.1&3150\\
    \hline
    *GDAS \citep{DongY19}&18.38&3.4&0.2\\
    *R-DARTS \citep{ZelaESMBH20}&18.01$\pm$0.26&-&1.6
    \\
              *DARTS$^{-}$ \citep{abs-2009-01027}&17.51$\pm$0.25&3.3&0.4\\
      *DropNAS \citep{HongL0TWL020} & 16.39&4.4&0.7 \\
\hline
\hline
     ${}^{\dag}$DARTS-1st \citep{liu2018darts}  &20.52$\pm$0.31 &1.8 &0.4\\
     $\;\;$LBT(RN18,DARTS-1st) (ours) &19.28$\pm$0.13& 1.9&0.5 \\
     $\;\;$LBT(RN50,DARTS-1st) (ours) &\textbf{18.74}$\pm$0.09&1.9&0.6 \\
     \hline
            *DARTS-2nd \citep{liu2018darts}  & 20.58$\pm$0.44&1.8&1.5 \\
             $\;\;$LBT(RN18,DARTS-2nd) (ours) &17.93$\pm$0.18 &2.0 &1.9 \\
               $\;\;$LBT(RN50,DARTS-2nd) (ours) &\textbf{17.06}$\pm$0.22 &2.1 &2.2  \\
        \hline
           *P-DARTS \citep{chen2019progressive}&17.49&3.6&0.3\\
       $\;\;$LBT(RN18,P-DARTS) (ours) &16.53$\pm$0.16&3.6&0.5 \\
         $\;\;$LBT(RN50,P-DARTS) (ours) &\textbf{16.29}$\pm$0.07&3.7&0.6 \\
                 \hline
          $\dag$PC-DARTS \citep{abs-1907-05737} &17.96$\pm$0.15&3.9&0.1 \\
       $\;\;$LBT(RN18,PC-DARTS)  (ours) &17.21$\pm$0.13& 3.8& 0.1\\
         $\;\;$LBT(RN50,PC-DARTS) (ours) &\textbf{16.88}$\pm$0.09&3.8& 0.2\\
        \bottomrule
    \end{tabular}
    \label{tab:cifar100}
\end{table} For example, ResNet with 50 layers (RN50) is generally considered to have stronger representation learning power than ResNet with 18 layers (RN18). In LBT applied to DARTS, P-DARTS, and PC-DARTS, we experimented with two student models with RN50 and RN18 as their architectures respectively. As can be seen, LBT with RN50 as the student achieves better performance than LBT with RN18 as the student. For example, on DARTS-2nd, LBT with RN50 achieves an error of 17.06\% while LBT with RN18 achieves an error of 17.93\%. As another example, on P-DARTS, the error achieved by LBT(RN50) is 16.29\%, which is  lower than the 16.53\% error achieved by LBT(RN18). The reason is that to teach a stronger student to learn better, the teacher has to be even stronger. For example, given a relatively weak student such as a CNN with only three layers, since there is a large room for the student to improve, an ordinary teacher such as a CNN with ten layers would be sufficient to teach the student to perform better. However, if the student (e.g., ResNet with 50 layers) is very strong whose performance is already high, it is very challenging to teach the student to improve unless the teacher is even stronger. To better train a strong student, the pseudo-labels generated by the teacher are required to be highly accurate. To achieve this goal, the teacher is forced to escalate the performance of its model to an excellent level. \textbf{Third}, our method LBT(RN50,P-DARTS)  achieves the lowest error among all methods in this table. This shows that our method is very competitive in bringing the NAS performance to a state-of-the-art level. 
\textbf{Fourth}, while our LBT framework can greatly help to improve the quality of searched architectures, it does not substantially increase the number of model parameters or search cost. As shown in the third column and fourth column, the model size and search cost of our methods are similar to those of baselines.

\begin{table}[t]
 \caption{
    Results on CIFAR-10,  including classification error (\%) on the test set, number of model weights (millions), and search cost (GPU days).
    * denotes that the results are taken from DARTS$^{-}$ \citep{abs-2009-01027}, NoisyDARTS \citep{abs-2005-03566},  and DrNAS \citep{abs-2006-10355}.
    The rest notations are the same as those in Table~\ref{tab:cifar100}.
    }
    \centering
    \begin{adjustbox}{width=0.73\columnwidth,center}
    \begin{tabular}{l|ccc}
    \toprule
    Method& Error(\%)& Param(M) & Cost\\
    \midrule
    *DenseNet
    \citep{HuangLMW17}&3.46&25.6 &-\\
    \hline
     *HierEvol \citep{liu2017hierarchical}&3.75$\pm$0.12& 15.7 &300\\
    *NAONet-WS \citep{LuoTQCL18} & 3.53 & 3.1&0.4 \\
        *PNAS \citep{LiuZNSHLFYHM18} &3.41$\pm$0.09  &3.2& 225\\
        *ENAS \citep{pham2018efficient} &2.89 & 4.6  &0.5 \\
    *NASNet-A \citep{zoph2018learning} & 2.65 & 3.3& 1800\\
    *AmoebaNet-B \citep{real2019regularized} & 2.55$\pm$0.05 & 2.8&3150  \\
    \hline
        *R-DARTS \citep{ZelaESMBH20} &2.95$\pm$0.21  &- & 1.6 \\
            *GDAS \citep{DongY19}&2.93& 3.4& 0.2 \\
    *SNAS \citep{xie2018snas} &2.85 & 2.8& 1.5\\
        *BayesNAS \citep{ZhouYWP19} &2.81$\pm$0.04 &3.4&0.2 \\
        *MergeNAS \citep{WangXYYHS20} &2.73$\pm$0.02 &2.9 & 0.2 \\
        *NoisyDARTS \citep{abs-2005-03566} &2.70$\pm$0.23&3.3  & 0.4 \\
            *ASAP \citep{NoyNRZDFGZ20} &2.68$\pm$0.11 & 2.5&0.2 \\
                *SDARTS
    \citep{abs-2002-05283}&2.61$\pm$0.02 & 3.3& 1.3 \\
            *DropNAS \citep{HongL0TWL020} &2.58$\pm$0.14 & 4.1&0.6 \\
            *PC-DARTS \citep{abs-1907-05737} &2.57$\pm$0.07&3.6& 0.1\\
    *FairDARTS \citep{abs-1911-12126} &2.54 &3.3 &0.4 \\
       *DrNAS \citep{abs-2006-10355} &2.54$\pm$0.03&4.0&  0.4\\ *GTN~\citep{abs-1912-07768}& 2.92$\pm$0.06 & 8.2&  0.67\\
       *GTN(F=128)~\citep{abs-1912-07768}& 2.42$\pm$0.03 & 97.9&  0.67\\
    \hline
        \hline
            *DARTS-1st \citep{liu2018darts} &3.00$\pm$0.14&3.3&  0.4\\
        $\;\;$LBT(RN18,DARTS-1st) (ours) &2.87$\pm$0.05 & 3.2& 0.6\\
             $\;\;$LBT(RN50,DARTS-1st) (ours) &\textbf{2.79}$\pm$0.07 &3.3 & 0.7\\
        \hline
               *DARTS-2nd \citep{liu2018darts} &2.76$\pm$0.09&3.3&  1.5\\
           $\;\;$LBT(RN18,DARTS-2nd) (ours)  & 2.65$\pm$0.03 &  3.4&  1.9 \\
            $\;\;$LBT(RN50,DARTS-2nd) (ours) &  \textbf{2.61}$\pm$0.05  &3.4 &2.1 \\
     \hline
    *P-DARTS \citep{chen2019progressive}& 2.50&3.4&  0.3\\
     $\;\;$LBT(RN18,P-DARTS) (ours)& 2.64$\pm$0.11& 3.4 & 0.4  \\
          $\;\;$LBT(RN50,P-DARTS) (ours)& 2.57$\pm$0.15& 3.4 & 0.5 \\
               \hline
      *PC-DARTS \citep{abs-1907-05737} &2.57$\pm$0.07&3.6& 0.1\\
     $\;\;$LBT(RN18,PC-DARTS) (ours)& 2.59$\pm$0.03& 3.7 & 0.1 \\
          $\;\;$LBT(RN50,PC-DARTS) (ours)& 2.56$\pm$0.04& 3.7 & 0.2 \\
        \bottomrule
    \end{tabular}
    \end{adjustbox}
    \label{tab:cifar10}
\end{table}

\begin{table*}[t]
  \caption{Results on ImageNet, including top-1 and top-5 classification errors on the test set, number of model weights (millions), and search cost (GPU days). * denotes that the results are taken from DARTS$^{-}$ \citep{abs-2009-01027} and DrNAS \citep{abs-2006-10355}. The rest notations are the same as those in Table~\ref{tab:cifar100}. 
    From top to bottom, on the first, second, and third panel are manually-designed networks, non-differentiable search methods, and differentiable search methods.  LBT(RN50,DARTS-2nd,CIFAR10) means the architecture is searched using LBT on CIFAR-10 with ResNet-50 as the student, where the search space is the same as that in DARTS-2nd. Similar meanings hold for other notations like this. 
    }
    \centering
    \begin{adjustbox}{width=\columnwidth,center}
    \begin{tabular}{l|cccc}
    \toprule
  \multirow{ 2}{*}{Method}   & Top-1  &Top-5 &Param & Cost \\
         & Error (\%) & Error (\%)&(M) & (GPU days)\\
    \midrule
    *Inception-v1 \citep{googlenet}&30.2 &10.1&6.6&- \\
    *MobileNet \citep{HowardZCKWWAA17} &  29.4& 10.5 &4.2&- \\
    *ShuffleNet 2$\times$ (v2) \citep{MaZZS18} &  25.1 &7.6 & 7.4&-\\
    \hline
    *NASNet-A \citep{zoph2018learning} &26.0 &8.4 &5.3 &1800 \\
    *PNAS \citep{LiuZNSHLFYHM18} &25.8 &8.1  &5.1 &225 \\
    *MnasNet-92 \citep{TanCPVSHL19} & 25.2 & 8.0& 4.4&1667\\
        *AmoebaNet-C \citep{real2019regularized} &  24.3 &7.6 &6.4&3150 \\
    \hline
     *SNAS-CIFAR10 \citep{xie2018snas} & 27.3 &9.2 &4.3 &1.5 \\
          *BayesNAS-CIFAR10 \citep{ZhouYWP19} &26.5 &8.9 &3.9&0.2 \\
                    *PARSEC-CIFAR10 \citep{abs-1902-05116} & 26.0 &8.4&5.6&1.0 \\
     *GDAS-CIFAR10 \citep{DongY19} &  26.0&8.5 &5.3 & 0.2\\
                 *DSNAS-ImageNet \citep{HuXZLSLL20} &25.7& 8.1 &- & -\\
          *SDARTS-ADV-CIFAR10 \citep{abs-2002-05283}&25.2& 7.8 &5.4& 1.3 \\
           *PC-DARTS-CIFAR10 \citep{abs-1907-05737} & 25.1 &7.8&5.3&0.1\\
                *ProxylessNAS-ImageNet \citep{cai2018proxylessnas} & 24.9 &7.5 &7.1 &8.3  \\
          *FairDARTS-CIFAR10 \citep{abs-1911-12126} &24.9 &7.5 &4.8 &0.4 \\
     *FairDARTS-ImageNet \citep{abs-1911-12126} &24.4 &7.4 &4.3 &3.0 \\
             *DrNAS-ImageNet \citep{abs-2006-10355} & 24.2 &7.3& 5.2&3.9\\
        *DARTS$^{-}$-ImageNet \citep{abs-2009-01027}&23.8& 7.0&4.9&4.5\\
     *DARTS$^{+}$-CIFAR100 \citep{abs-1909-06035}&23.7& 7.2&5.1&0.2\\
     \hline
       \hline
            *DARTS-2nd(CIFAR10) \citep{liu2018darts}  & 26.7 &8.7&4.7&1.5 \\
        $\;\;$LBT(RN50,DARTS-2nd,CIFAR10) (ours) & \textbf{25.5}  &\textbf{7.9} & 4.8 & 2.1 \\
        \hline
          *P-DARTS(CIFAR10) \citep{chen2019progressive}&24.4 &7.4&4.9&0.3\\
        $\;\;$LBT(RN50,P-DARTS,CIFAR10) (ours) & \textbf{24.1} &\textbf{7.2}  &4.9 &  0.5\\
        \hline
             *P-DARTS(CIFAR100) \citep{chen2019progressive}&24.7& 7.5&5.1&0.3\\
           $\;\;$LBT(RN50,P-DARTS,CIFAR100) (ours) & \textbf{24.2} & \textbf{7.1} &5.3 &0.6
           \\
           \hline
            *PC-DARTS(ImageNet) \citep{abs-1907-05737} &  24.2 &7.3&5.3&3.8\\
 $\;\;$LBT(RN50,PC-DARTS,ImageNet) (ours) &\textbf{23.5}  & \textbf{6.8} &5.4 & 4.1
           \\
        \bottomrule
    \end{tabular}
    \end{adjustbox}
    \label{tab:imagenet}
\end{table*}

In Table~\ref{tab:cifar10}, we show the results on CIFAR-10, including the classification error on the test set, number of model parameters, and search cost measured by GPU days. From this table, we make similar observations as those in Table~\ref{tab:cifar100}. \textbf{First}, our proposed LBT framework is widely effective in helping different NAS methods to improve the quality of searched architectures. For example, applying LBT to DARTS-2nd reduces the error from 2.76\% to 2.61\%. This further demonstrates that by teaching a student model to learn well, the teacher can improve itself greatly. In the GTN~\citep{abs-1912-07768}  baseline approach where the architecture of a student is searched by leveraging the synthetic data generated by a trainable generative model, the classification error is 2.92\%, which is much worse than those achieved by our methods, while the number of parameters in GTN is twice  more than ours. Setting the network width to 128, GTN achieves an error of 2.42\%; however, the resulting number of parameters in GTN is about 30 times more than ours. GTN focuses on searching the student's architecture while our method focuses on searching the teacher's architecture. These results demonstrate that searching  teacher's architecture is more advantageous than  searching  student's architecture. 
\textbf{Second}, using ResNet with 50 layers (RN50) as the student results in better performance than using ResNet-18 (RN18), which further demonstrates that teaching a stronger student can drive the teacher to learn better. \textbf{Third}, while achieving better classification accuracy than baselines, our method does not substantially increase the model size or search cost compared with baselines.

The results on ImageNet are shown in Table~\ref{tab:imagenet}, including top-1 and top-5 classification errors on the test set, number of weight parameters (millions), and search costs (GPU days). LBT(RN50,DARTS-2nd,CIFAR10), which is the architecture searched by applying LBT to DARTS-2nd on CIFAR10 with ResNet-50 as the student, achieves lower error than DARTS-2nd(CIFAR10) which does not use LBT. Similarly, LBT(RN50,P-DARTS,CIFAR10) outperforms P-DARTS(CIFAR10), LBT(RN50,P-DARTS,CIFAR100)  outperforms P-DARTS(CIFAR100), and LBT(RN50,PC-DARTS,ImageNet) outperforms PC-DARTS(ImageNet). These results again demonstrate the effectiveness of our method in improving a model by letting it teach another model to learn well.

\subsection{Ablation Studies}
In this section, we perform several ablation studies to better understand the individual learning stages in LBT. For each ablation setting, we compare it with the full LBT framework.

\begin{itemize}[leftmargin=*]
    \item \textbf{Ablation setting 1}. In this setting, the teacher updates its architecture by minimizing the validation loss of the student only, without considering the validation loss of itself. 
     The corresponding formulation is: \begin{equation*}
\begin{array}{l}
\underset{A}{\textrm{min}}
  \;\;   L(S^*(T^*(A)), D_s^{(\textrm{val})})\\
      s.t. \;\;\; S^*(T^*(A)) =
      \underset{S}{\textrm{min}}
    \;\;  L(S, D_s^{(\textrm{tr})} )+\lambda  L(S, D_{pl}(D_u,T^*(A)))\\
    \quad\quad T^*(A) =
    \underset{T}{\textrm{min}} \; L(A, T,D_t^{(\textrm{tr})})
\end{array}
\label{eq:ab-1}
\end{equation*}
In this study, $\lambda$ is set to 1. The student's architecture is  ResNet-18.  LBT is applied to DARTS-2nd. 
  \item \textbf{Ablation setting 2}. In this setting, in the second stage of LBT, only the pseudo-labeled dataset is used to train the student; $D_s^{(\textrm{tr})}$ labeled by humans is not used. 
  The corresponding formulation is: 
  \begin{equation*}
\begin{array}{l}
\underset{A}{\textrm{min}}
  \;\;  L(T^*(A),A,D_t^{(\textrm{val})})) + \gamma  L(S^*(T^*(A)), D_s^{(\textrm{val})})\\
      s.t. \;\;\; S^*(T^*(A)) =
      \underset{S}{\textrm{min}}
    \;\;   L(S, D_{pl}(D_u,T^*(A)))\\
    \quad\quad T^*(A) =
    \underset{T}{\textrm{min}} \; L(A, T,D_t^{(\textrm{tr})})
\end{array}
\label{eq:ab-2}
\end{equation*}
In this study,   $\gamma$ is  set to 1. The student's architecture is ResNet-18. LBT is applied to DARTS-2nd.
    \item Ablation study on $\lambda$. We investigate how the teacher's test error changes with the tradeoff parameter $\lambda$ in Eq.(\ref{eq:lbt}). 
   In this study, the other tradeoff parameter $\gamma$ in Eq.(\ref{eq:lbt}) is set to 1. For either CIFAR-100 or CIFAR-10, from their training set and validation set, 5K examples are uniformly sampled to form a new test set. Architecture search is performed on the remaining training and validation sets. Architecture evaluation result is reported on the 5K new test set. 
  Student's architecture is ResNet-18. LBT is applied to DARTS-2nd.
    \item Ablation study on $\gamma$. We investigate how the teacher's test error changes with the tradeoff parameter $\gamma$ in Eq.(\ref{eq:lbt}). 
   In this study, the other tradeoff parameter $\lambda$  is set to 1. Similar to the ablation study on $\lambda$, the test error is reported on the 5K dataset. 
  Student's architecture is ResNet-18. LBT is applied to DARTS-2nd.
\end{itemize}

In Table~\ref{tab:ab1}, we show the classification errors on the test set of CIFAR-10 and CIFAR-100 in ablation setting 1. As can be seen, on both datasets, minimizing both student's validation loss and teacher's validation loss results in better architectures for the teacher than minimizing student's validation only. The reason is that student's validation loss indirectly measures the quality of the teacher's architecture. How well the student performs depends on not only how well the teacher teaches the student but also how strong the student itself is. If the student is a very strong learner, its validation loss may be largely determined by the student itself and less influenced by the teacher. In this case, student's validation would be a relatively weak signal for guiding the learning of the teacher. In contrast, the validation loss of the teacher directly depends on its architecture and can serve as a direct (hence strong) signal to guide the teacher to learn. In the end, combining the direct signal (teacher's validation loss) and indirect signal (student's validation loss) together is more beneficial than using the indirect signal only.  

\begin{table}[t]
\caption{Classification errors in  ablation setting 1. ``Student only" 
    means that only the student's validation loss is used to update the teacher's architecture. 
    ``Student + teacher" means that both the student's validation loss and teacher's validation loss are minimized to update the teacher's architecture. 
    }
        \centering
    \begin{tabular}{l|c}
    \hline
    Method & Error (\%)\\
    \hline
    Student only (CIFAR-100) &  20.27$\pm$0.24 \\ 
            Student + Teacher (CIFAR-100) &\textbf{17.93}$\pm$0.18  \\
         \hline
             Student only (CIFAR-10) & 3.01$\pm$0.08  \\
         Student + teacher (CIFAR-10) &\textbf{2.61}$\pm$0.05   \\
         \hline
    \end{tabular}
    \label{tab:ab1}
\end{table}

\begin{table}[t]
\caption{
    Classification errors in ablation setting 2. ``Pseudo labels only" means in the second learning stage, only the pseudo-labeled dataset is used to train the student. ``Pseudo labels + human labels" means both the pseudo-labeled dataset and a human-labeled dataset $D_s^{(\textrm{tr})}$ are used to train the student. 
    }
    \centering
    \begin{tabular}{l|c}
    \hline
    Method & Error (\%)\\
    \hline
         Pseudo labels only (CIFAR-100) & 19.82$\pm$0.31 \\ 
           Pseudo labels + human labels (CIFAR-100) &\textbf{17.93}$\pm$0.18  \\
         \hline
          Pseudo labels only (CIFAR-10) & 2.93$\pm$0.07  \\
         Pseudo labels + human labels (CIFAR-10) &\textbf{2.61}$\pm$0.05  \\
         \hline
    \end{tabular}
    \label{tab:ab4}
\end{table}

In Table~\ref{tab:ab4}, we show the classification errors on the test sets of CIFAR-10 and CIFAR-100 in ablation setting 2. We can see that using both the pseudo-labeled dataset and human-labeled dataset to train the student yields better performance than using the pseudo-labeled dataset only. The reason is that since the pseudo-labels are automatically generated by a model, they are not entirely reliable. Trained on less reliable labels, the student's model may have low quality and a poorly-performing student cannot drive the teacher to learn better. This risk can be reduced by incorporating human-provided labels which are more reliable. As a result, using human labels and pseudo-labels jointly yields better performance than solely using pseudo-labels.

\begin{figure}[t]
    \centering
 \includegraphics[width=0.49\columnwidth]{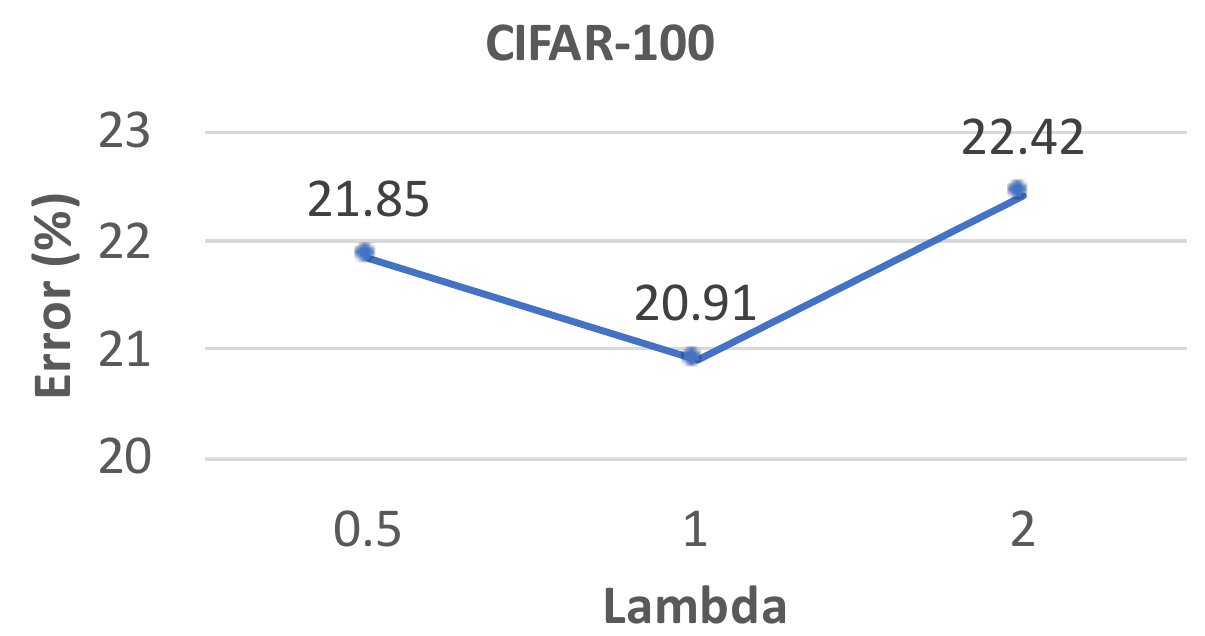}
  \includegraphics[width=0.49\columnwidth]{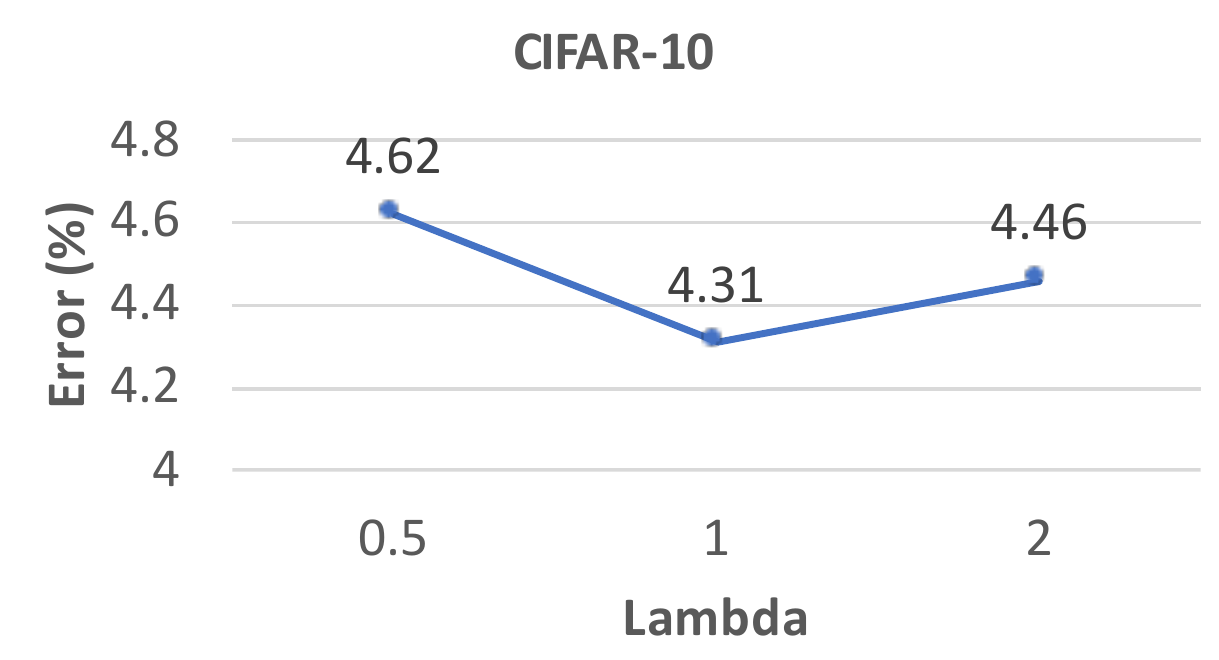}
       \caption{How errors change as $\lambda$ increases.}
 \label{fig:lambda}
\end{figure}

\begin{figure}[t]
    \centering
 \includegraphics[width=0.49\columnwidth]{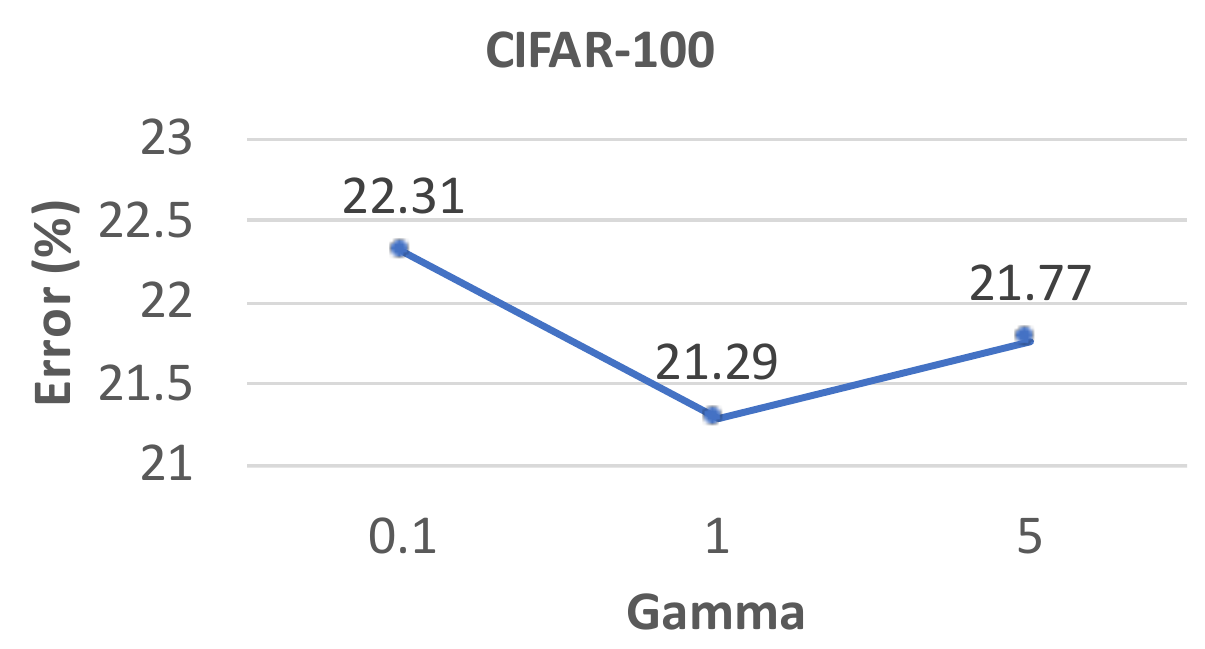}
  \includegraphics[width=0.49\columnwidth]{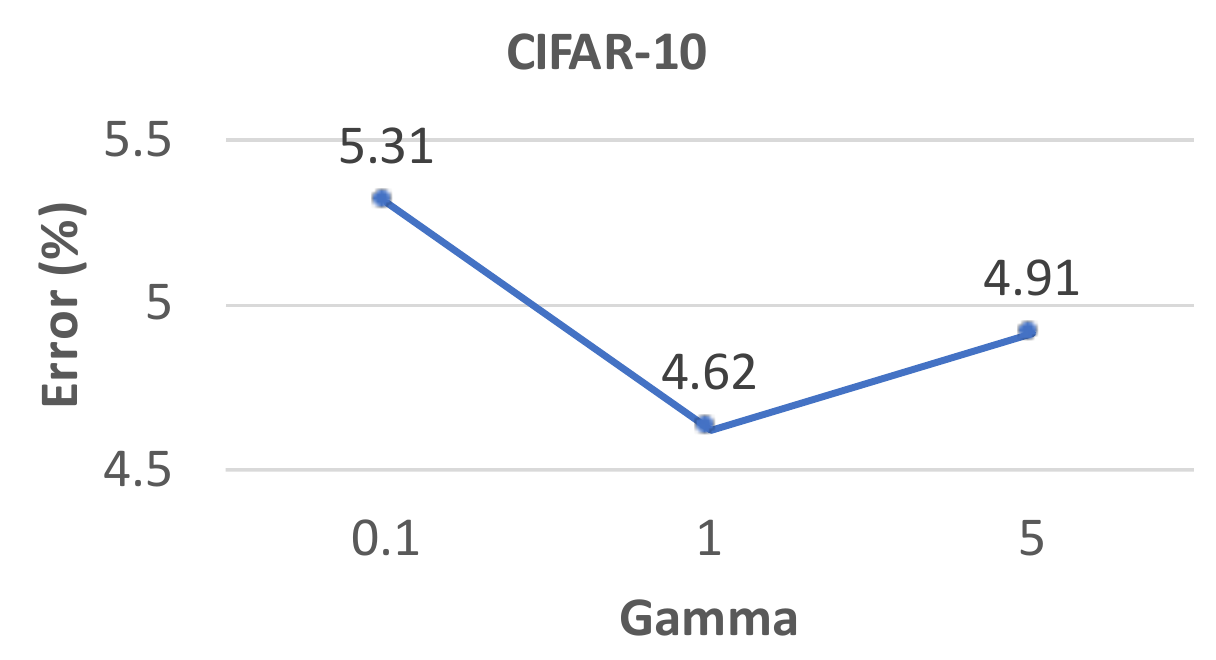}
       \caption{How errors change as $\gamma$ increases.}
 \label{fig:gamma}
\end{figure}

In Figure~\ref{fig:lambda}, we show how the classification errors of LBT on CIFAR-10 and CIFAR-100 vary as $\lambda$ increases. 
From the plot on CIFAR-100, we observe the following. First, when $\lambda$ increases from 0.5 to 1, the error decreases. This is because a larger $\lambda$ incurs a stronger effect of teaching, where the training of the student relies more on the pseudo-labeled dataset created by the teacher. When the teaching effect is strong, the teacher can gain more feedback from the student's performance, which helps the teacher to learn better. Second, if we continue to increase $\lambda$, the performance becomes worse. The reason is that if $\lambda$ is too large, the teaching effect would be excessively strong. Under such circumstances, the student is mainly trained on the pseudo labels which are less reliable than human-provided labels and consequently its model may be of low quality. A mediocre student will not be very helpful in driving the teacher to improve. Similar phenomenon are observed from the plot on CIFAR-10.

In Figure~\ref{fig:gamma}, we show how the classification errors of LBT vary as we increase $\gamma$. As can be seen, on CIFAR-100, when we increase $\gamma$ from 0.1 to 1, the error decreases. This is because a larger $\gamma$ encourages the teacher to pay more attention to the feedback obtained from the student. This feedback is valuable because the validation performance of the student reflects the correctness of the pseudo-labels generated by the teacher and the quality of pseudo-labels reflects the quality of the teacher's architecture. Paying more attention to such feedback enables the teacher to identify its weakness and strive for improvement. However, if $\gamma$ is too large, the learning of the teacher's architecture would be guided excessively by student's validation loss which is an indirect (hence weaker signal) but inadequately influenced the validation loss of the teacher itself which is a direct (hence stronger signal). Similar observations can be made from the plot on CIFAR-10 as well.

\section{Conclusions}
Motivated by the teaching-driven learning methodology of humans, we propose a novel machine learning framework called learning by teaching (LBT). LBT improves the training of a teacher model by encouraging the teacher to teach what it has learned to a student model. Based on how the student performs on a validation set, the teacher refines its model. The intuition behind LBT is that  to be able to teach others to well accomplish a task, the teacher itself needs to thoroughly understand this task and know how to perform it excellently. We develop a multi-level optimization framework to formulate LBT and the framework consists of three learning stages:  the teacher learns; the teacher teaches the student by pseudo-labeling; the teacher improves its architecture based on the validation results of itself and of the student. Our framework is applied for neural architecture search  on CIFAR-100, CIFAR-10, and ImageNet. The results demonstrate the effectiveness of our method.

\bibliography{release-2}

\end{document}